\documentclass{article} 
\usepackage{iclr2024_conference,times}

\usepackage{amsmath,amsfonts,bm}

\def\eqref#1{equation~\ref{#1}}

\def\1{\bm{1}}

\DeclareMathAlphabet{\mathsfit}{\encodingdefault}{\sfdefault}{m}{sl}
\SetMathAlphabet{\mathsfit}{bold}{\encodingdefault}{\sfdefault}{bx}{n}

\usepackage[hidelinks]{hyperref}
\usepackage{url}
\usepackage{amsmath}
\usepackage{booktabs}       
\usepackage{amsfonts}       
\usepackage{xcolor}         
\usepackage{graphicx}
\usepackage{amssymb}
\usepackage{multirow}
\usepackage{algpseudocode}
\usepackage{bbm}
\newcommand{\fullname}{Mixture-of-Modules}
\newcommand{\name}{MoM}

\title{\fullname: \\Reinventing Transformers as Dynamic Assemblies of Modules}

\author{Zhuocheng Gong\textsuperscript{1}\footnotemark[1]\quad Ang Lv\textsuperscript{2}\footnotemark[1]\quad Jian Guan\textsuperscript{3,4}\footnotemark[2]\quad Junxi Yan\textsuperscript{3}\quad Wei Wu\textsuperscript{4}\quad Huishuai Zhang\textsuperscript{1}\\ 
\textbf{Minlie Huang\textsuperscript{3}\quad Dongyan Zhao\textsuperscript{1}\footnotemark[2]\quad Rui Yan\textsuperscript{2}\footnotemark[2]}\\
\textsuperscript{1}Peking University \quad 
\textsuperscript{2}Renmin University \quad 
\textsuperscript{3}Tsinghua University \quad 
\textsuperscript{4}Ant Group \\
\texttt{\{gzhch,zhanghuishuai,zhaody\}@pku.edu.cn}, \texttt{\{anglv,ruiyan\}@ruc.edu.cn}\\
\texttt{\{j-guan19,yanjx21\}@mails.tsinghua.edu.cn}, \texttt{aihuang@tsinghua.edu.cn}\\
\texttt{congyue.ww@antgroup.com}\\
}

\iclrfinalcopy 
\begin{document}

\maketitle

\renewcommand{\thefootnote}{\fnsymbol{footnote}}
\footnotetext[1]{Equal Contributions.}
\footnotetext[2]{Corresponding authors.}
\renewcommand*{\thefootnote}{\arabic{footnote}}

\begin{abstract}
Is it always necessary to compute tokens from shallow to deep layers in Transformers? The continued success of vanilla Transformers and their variants suggests an undoubted ``yes’’. In this work, however, we attempt to break the depth-ordered convention by proposing a novel architecture dubbed mixture-of-modules (MoM), which is motivated by an intuition that any layer, regardless of its position, can be used to compute a token as long as it possesses the needed processing capabilities.
The construction of MoM starts from a finite set of modules defined by multi-head attention and feed-forward networks, each distinguished by its unique parameterization. Two routers then iteratively select attention modules and feed-forward modules from the set to process a token. The selection dynamically expands the computation graph in the forward pass of the token, culminating in an assembly of modules. 
We show that MoM provides not only a unified framework for Transformers and their numerous variants but also a flexible and learnable approach for reducing redundancy in Transformer parameterization.
We pre-train various \name s using OpenWebText. Empirical results demonstrate that \name s, of different parameter counts, consistently outperform vanilla transformers on both GLUE and XSUM benchmarks. 
More interestingly, with a fixed parameter budget, MoM-large enables an over 38\% increase in depth for computation graphs compared to GPT-2-large, resulting in absolute gains of 1.4 on GLUE and 1 on XSUM.  On the other hand, MoM-large also enables an over 60\% reduction in depth while involving more modules per layer, yielding a 16\% reduction in TFLOPs and a 43\% decrease in memory usage compared to GPT-2-large, while maintaining comparable performance. \footnote{Code is available at \href{https://github.com/gzhch/MoM}{https://github.com/gzhch/MoM}}

\end{abstract}

\section{Introduction}

Transformer-based language models~\citep{vaswani} have demonstrated remarkable abilities across a wide range of challenging natural language tasks~\citep{bubeck2023sparks}. 
In addition, the success of Transformer in natural language processing (NLP) is also inspiring innovations in other fields such as computer vision~\citep{Peebles_2023_ICCV, agostinelli2023musiclm} and biomedicine~\citep{medical, protein}. 
A Transformer architecture typically consists of stacked layers that are identical in structure, whereby layers are organized in the order of depth, using the output of the previous layer as the input for the next. While this design convention has been widely accepted as a matter of course in the Transformer era, we challenge it by reconsidering whether the static and depth-ordered organization can fully unleash the potential of Transformers,  given the well-known issues of over-parameterization~\citep{zeng2023learning} and efficiency~\citep{mod}.

Before us, some rudimentary studies have touched on the question--they dissect Transformer into modules such as attention heads and feed-forward networks (\texttt{FFN}s) and allow relatively flexible module call order. 
For example, Mixture-of-Experts (MoE)~\citep{shazeer2017}) sets up multiple \texttt{FFN}s within the same layer and activates a specific subset during inference. 
Early-exiting~\citep{PABEE,xin-etal-2020-deebert,CALM_2022} and Mixture-of-Depths (MoD)~\citep{mod}) bypass certain layers when computing each token. On the one hand, these efforts indeed lead to improvements in terms of either efficacy or efficiency through the introduction of dynamic mechanisms into the vanilla structure of Transformers, and thus corroborate our questioning regarding the established convention; on the other hand, they still follow the depth-ordered paradigm (i.e., tokens are passed from shallow layers to deep layers), leaving significant room for better architectures. 

\begin{figure*}[t]
    \centering
    \includegraphics[width=0.9\linewidth]{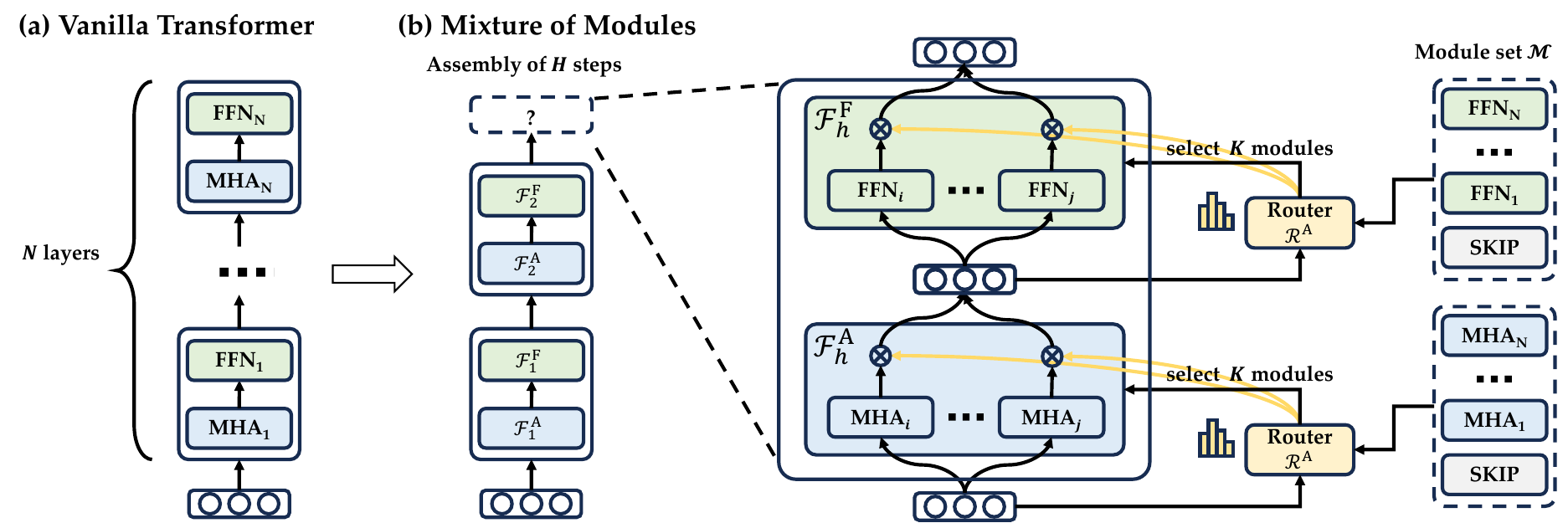}
    \caption{\fullname\ reinvents Transformers as dynamic assemblies of modules.
    In (b), we illustrate the ongoing construction of an \name\ model during the forward computation. 
    The assembly lasts $H$ rounds, with the current illustration showcasing progress in the third round.
    For each token, routers select the best $K$ attention modules, denoted as $m^{\text{A}}_{k}$, and the best $K$ feed-forward network modules, denoted as $m^{\text{F}}_{k}$, from a module set $\mathcal{M}$ (including ``\texttt{SKIP}'' modules).
    These selected modules collectively constitute assembled modules $\mathcal{F}^{\text{A}}$ and $\mathcal{F}^{\text{F}}$, which are then appended to the existing computation graph. Detailed notations are presented in \S\ref{sec:method}.}
    \label{fig:introduction}
    \vspace{-5mm}
\end{figure*}

In this work, we completely disrupt the traditional practice in the design of Transformers by breaking down the depth-ordered organization. Numerous studies have indicated that knowledge in Transformers is often dispersed across multiple \texttt{FFN}s in different layers~\citep{geva-etal-2021-transformer,mcgrath2023hydra,lv2024interpreting}, and many attention heads serve similar or identical functions, such as copying specific token information towards the end position of the input~\citep{olsson2022context,wang2023interpretability}. 
Encouraged by this evidence, we pose the question of whether the computation of a token can ``move'' freely across layers, that is the token can be computed by flowing to modules in deeper layers, sticking to modules of the same layer, or even going back to modules in previous layers.
To answer the question, we propose a novel architecture dubbed \fullname~(\name) in which the core idea is to define a neural network as dynamic assemblies of modules derived from vanilla Transformer, as depicted in Figure~\ref{fig:introduction}.

The basis of \name\ is a finite set of modules. 
Each module is defined by multi-head attention (\texttt{MHA}), a feed-forward network (\texttt{FFN}) (including Add \& Norm), or a specialized module labeled ``\texttt{SKIP}''.
Each \texttt{MHA} or \texttt{FFN} module is identical in structure and different in parameterization. 
\texttt{SKIP} enables skip operations for arbitrary tokens at arbitrary time steps. 
Given a token, each time two routers select modules from the set and integrate the modules into the computation graph during the forward pass. 
Hence, the whole computation graph of the token is formed as an assembly of modules, and the routers learn to optimize the organization of the modules in the assembly.
We introduce a two-phase approach for training \name\ models. In the first phase, we pre-train a vanilla Transformer on a large-scale corpus. Then, in the second phase, we decompose the pre-trained Transformer into modules as a warm-up of \name, randomly initialize the routers, and further update both the modules and the routers under the mechanism of dynamic assembly. By this means, we can both enhance parameter utilization and accelerate the convergence of the model.

\name\ has three major advantages over existing Transformer-based architectures: (1) it provides a unified framework for various Transformer variants, incorporating popular methods such as mixture-of-experts, early-exiting, and mixture-of-depths as special cases. 
The framework sheds light on architecture design for future works; (2) it brings unprecedented flexibility in forward computation. With the dynamic assembly mechanism, ``depth'' and ``parameter count'' are no longer entangled as they are in the conventional sense. One can build powerful architectures by either enlarging the module pool (i.e., increasing parameter count) or increasing the depth (parameter count can be fixed). 
Hence, \name\ offers a dynamic and learnable approach to reducing redundant parameters in Transformers; and (3) it offers efficient structures that achieve performance comparable to vanilla Transformers but require significantly fewer  FLOPs and less memory in forward computation.

We pre-train \name\ in three sizes--122M (small), 346M (medium), and 774M (large)-- using OpenWebText~\citep{Gokaslan2019OpenWeb}, and assess their performance with GLUE~\citep{wang2018glue} and XSUM~\citep{narayan2018don}. Empirical results indicate that (1) \name s, across all the three parameter counts, consistently outperform vanilla GPT-2 models on both text understanding and generation tasks;
(2) parameters are quite redundant in vanilla Transformers. One can develop an \name\ that is at least 30\% deeper than a vanilla GPT-2, resulting in at least 1.4 absolute gain in GLUE and at least 1 absolute gain in XSUM. Moreover, one can further remove $50$\% of the \texttt{MHA} modules and $25$\% of the \texttt{FFN} modules from the module pool of the model, while maintain comparable performance\footnote{After removing $50$\% of the \texttt{MHA} modules and $25$\% of the \texttt{FFN} modules, the validation loss increased by $0.065$, the GLUE average score dropped from $81.98$ to $81.34$, and the ROUGE average in XSUM dropped from $19.34$ to $18.55$.}; 
and (3) for those concerned with efficiency, \name-large can reduce TFLOPs by 16\% and memory usage by 42\% in forward computation, while maintaining comparable performance to GPT-2-large, via properly increasing the number of modules and compressing the model depth.

Our contributions are three-fold:  (1) proposal of \fullname\ to disrupt the depth-ordered convention in Transformer construction, and reinvent Transformers as dynamic assemblies of modules; (2) empirical verification of the efficacy of \fullname\ on GLUE and XSUM; and (3) a series of new insights into the over-parameterization issue of vanilla Transformers, and their implications for future architecture design. The code for implementing \name\ is open-sourced at \ref{??}.






\section{Related works}
\label{sec:related}

\name\ owns a dynamic mechanism of module selection and combination, and thus is related to conditional computation techniques~\citep{bengio2013estimating,davis2014lowrank,cho2014exponentially}. 
Existing work on conditional computation can be categorized into two groups: \emph{dynamic depth} and \emph{dynamic width}. In these fields, terms such as gating and routing are used interchangeably, hereafter referred to as ``routers'' for clarity in presentation.

As a typical approach in dynamic depth, \textit{Early-exiting} ~\citep{graves2017adaptive,figurnov2017spatially,CALM_2022} accelerates model inference through terminating forward computation at intermediate layers. The decision to exit often relies on confidence-based metrics~\citep{EETFMR_2020,varshney2023accelerating,xin-etal-2020-deebert} or pre-determined strategies~\citep{Liu2020FasterDT,delcorro2023skipdecode}.  With some degree of generalization, \textit{Layer-skip}~\citep{srivastava2015highway,SkipNet,bapna2020controlling} represents a more adaptive variant of early-exiting, enabling certain layers to be skipped without terminating the entire forward computation. Existing works mainly facilitate it by training a router~\citep{zeng2023learning,mod} or layer pruning~\citep{Yang2024LaCoLL,kim2024shortened}. Finally, if we view parameter copying as a particular way to increase network depth with controlled model size, then some parameter sharing methods~\citep{dehghani2018universal,Lan2020ALBERT}, wherein certain modules or layers share parameters, also fall in the dynamic depth group. 

In terms of dynamic width, \textit{Mixture-of-Experts} (MoE,~\citep{shazeer2017,lepikhin2021gshard,switch-trans}) is a representative method.
An MoE model conceptualizes an FFN module as an ``expert'' for storing knowledge.
Comprising multiple such experts, an MoE layer replaces the traditional FFN layer within Transformers, aiming for superior performance in handling knowledge-related tasks.
During forward computation, a router network dynamically assigns each token to the top $K$ experts out of a total of $N$ experts, thereby increasing the maximum network width by $K$ times. Other dynamic width methods, such as CODA~\citep{lei2024conditional} and CoLT5~\citep{ainslie2023colt5}, use similar routing mechanism to select whether a token passes through a heavy or light pathway for not only each FFN layer but also each attention layer.

\name\ breaks the depth-ordered paradigm followed by existing approaches when performing forward computation. It not only unifies a number of approaches described above but also offers a more flexible and learnable way to achieve conditional computation. 

\section{Methodology}
\label{sec:method}
The idea of \fullname\ (\name) is inspired by the theory presented in ``the society of mind'' by Marvin Minsky~\citep{minsky1986society}, which explains the true intelligence as certain and very special ways of combinations of simple and modular units (in the book, they are termed ``agents''). 
In \S\ref{sec:mom}, we first provide an overview of \name. Then, we detail the assembly of modules and the routers in \S\ref{sec:dynamic-construct} and \S\ref{sec:router}. 
After that, we present the training procedure of \name\ in \S\ref{sec:optimize}. 
Finally in \S\ref{sec:unify}, we show that \name\ unifies various techniques of dynamic computation allocation within Transformers as special cases. 

\subsection{\fullname\ (\name)}
\label{sec:mom}

Before delving into the details, we first give a brief description of the workflow of \name. \name\ views the construction of an $H$-depth transformer as an $H$-step iterative assembly process. In each assembly step, router $\mathcal{R}$ dynamically selects $K$ modules from a module set $\mathcal{M}$ for each token. Then these selected modules are assembled guided by the assembling function $\phi$.
Formally, \name\ can be defined by a 5-tuple $<\mathcal{M}, \mathcal{R}, \phi, K, H>$. 

$\mathcal{M}$ is the set that contains all possible modules, where modules are defined as atomic units that could be assembled. 
There are two types of modules in a Transformer model, \emph{i.e.}, the multi-head self-attention module (\texttt{MHA}) and the feed-forward network module (\texttt{FFN}), denoted as $m^{\text{A}}$ and $m^{\text{F}}$, respectively. 
In addition, denote the input hidden state of a Transformer layer as $\mathbf{x} \in \mathbb{R}^{d}$, we include a special module $m^{\text{S}}: \mathbf{x}\mapsto\mathbf{x}$ in $\mathcal{M}$, which means the absence of an operation applied to the token, allowing for skipping one round of computation. 
Therefore, in \name, we have:
\begin{equation}
\begin{aligned}
    \mathcal{M} = \{m^{\text{A}}_i\}_{i=1}^{N_{\text{A}}} \cup \{m^{\text{F}}_i\}_{i=1}^{N_{\text{F}}} \cup \{m^{\text{S}}\},
\end{aligned}
\end{equation}
where $N_{\text{A}}$ are $N_{\text{F}}$ refer to numbers of \texttt{MHA}s and \texttt{FFN}s, respectively.

$\mathcal{R}$ is a router responsible for dynamically selecting appropriate modules from $\mathcal{M}$ and assembling them into the computation graph. 
We use distinct routers for \texttt{MHA}s and \texttt{FFN}s, denoted as $\mathcal{R}^{\text{A}}$ and $\mathcal{R}^{\text{F}}$ respectively. 
The output of the router $\mathcal{R}^{\mathcal{X}}$ is an $(N_{\mathcal{X}}+1)$-dimensional distribution wherein each item represents the weight assigned to each module $m^{\mathcal{X}}_i$ as well as $m^{\text{S}}$.
Formally:
\begin{equation}
\begin{aligned}
    &\mathcal{R}^{\mathcal{X}}:\quad \mathbf{x}\mapsto\mathbf{r}^{\mathcal{X}}, \\
    &\mathbf{x}\in\mathbb{R}^d,\quad \mathbf{r}^{\mathcal{X}}\in\mathbb{R}^{N_{\mathcal{X}}+1}, \quad \mathcal{X}\in\{\text{A},\text{F}\}.
\end{aligned}
\end{equation}

An \name\ model is dynamically assembled step by step. In each step, based on the output of $\mathcal{R}_{\mathcal{X}}$, $K_{\mathcal{X}}$ modules are selected from $\mathcal{M}$ and assembled together. The assembly process lasts $H$ steps, as detailed in \S\ref{sec:dynamic-construct}.
Further elaboration on these routers, including their architecture and the working pipeline, is deferred to \S\ref{sec:router}.
Notably, in \name, dynamic assembly occurs at the token level, wherein each token is independently and dynamically assigned by routers to appropriate modules for processing. 

\subsection{Dynamic assembly of modules}
\label{sec:dynamic-construct}
We delve into how an \name\ model is dynamically assembled.
The construction is an iterative process where in the $h$-th step (i.e., the $h$-th layer of the model being constructed), we have the input $\mathbf{x}_h$. 
The subscript $h$ is omitted when there's no ambiguity.
The router $\mathcal{R}$ selects $K$ modules with the largest routing weight.
We denote the indices of the selected modules as $\mathcal{K}_{\mathcal{X}}=\{i|r_i\in\text{TopK}(\mathbf{r}^{\mathcal{X}})\}$.
Then the selected modules are assembled together through the assembling function $\phi$. Formally,
\begin{equation}
\begin{aligned}
    \phi:\quad <\mathcal{M},\mathcal{R}^{\mathcal{X}}, \mathbf{x}_{h}>\ \mapsto\ \mathcal{F}^{\mathcal{X}}_h,\quad\mathcal{X}\in\{\text{A},\text{F}\},
\end{aligned}
\end{equation}
where $\mathcal{F}^{\mathcal{X}}_h$ represents the assembled modules.
These assembled modules transform the input $\mathbf{x}_h$ into the output $\mathbf{x}_{h+1}$.
We hope the role of the $h$-th step of \name\ assembly is somewhat akin to the $h$-th Transformer block in the conventional sense. 
Therefore, we establish two rounds of routing and assembling in each assembly step: one for \texttt{MHA} and the other for \texttt{FFN}. 
The forward computation of \name\ models at the $h$-th step assembly can be represented as:
\begin{equation}
\begin{aligned}
    &\mathbf{u}_{h} = \mathcal{F}^{\text{A}}_h(\mathbf{x}_{h}) + \mathbf{x}_{h},\\
    &\mathbf{x}_{h+1} = \mathcal{F}^{\text{F}}_h(\mathbf{u}_{h}) + \mathbf{u}_{h}.
\end{aligned}
\end{equation}
We employ Pre-norm in \name, which normalizes the input before feeding to assembled modules $\mathcal{F}^{\mathcal{X}}$. 
The dynamic assembly process is depicted in Figure~\ref{fig:introduction}(b).
We now introduce the detailed formalization for $\mathcal{F}^{\text{A}}$ and $\mathcal{F}^{\text{F}}$, respectively.

\textbf{Assembly of attention modules} ($\mathcal{F}^{\text{A}}$). \quad
We begin by considering the scenario where the $m^{\text{S}}$ module is not selected by routers. 
Suppose that an \texttt{MHA} module contains $Z$ individual heads, then 
the assembly of $K_{\text{A}}$ \texttt{MHA} modules (i.e., the computation process of $\mathbf{o} = \mathcal{F}^{\text{A}}(\mathbf{x})$) is defined as:

\begin{equation}
    \begin{aligned}
        &\mathbf{o} = \mathbf{a}\sum_{k\in\mathcal{K}_{\text{A}}}r^{\text{A}}_{k}\cdot \mathbf{W}_{k}^O,\\ 
        &\mathbf{r}^{\text{A}} = \mathcal{R}^{\text{A}}(\mathbf{x}) =\left(r^{\text{A}}_{1},\ldots, r^{\text{A}}_{k}, \ldots, r^{\text{A}}_{K_A}\right), \\        
        &\mathbf{a} = \left(\mathbf{x}\sum_{k\in\mathcal{K}_{\text{A}}}\mathbf{W}_{k,z}^V\right)\cdot \\
        &\texttt{softmax}\left(\frac{(\mathbf{X}\sum_{k\in\mathcal{K}_{\text{A}}}\mathbf{W}_{k,z}^Q)(\mathbf{X}\sum_{k\in\mathcal{K}_{\text{A}}}\mathbf{W}_{k,z}^K)^{\top}}{\sqrt{d_{\text{head}}}}\right),\\
    \end{aligned}
\end{equation}

where $\mathbf{X}\in\mathbb R^{L\times d}$ is the input representation of the sequence, 
$\mathbf{W}_{k,z}^Q, \mathbf{W}_{k,z}^K, \mathbf{W}_{k,z}^V \in \mathbb{R}^{d\times d_{\text{head}}}$, and $\mathbf{W}_{k}^O \in \mathbb{R}^{d_{\text{head}} \times d}$ are weight matrices with $d_{\text{head}} = d / Z$. When the $m^{\text{S}}$ module is selected, the operation of $\mathcal{F}^{\text{A}}$ only involves the remaining $K_{\text{A}} - 1$ attention modules.

\textbf{Assembly of feed-forward networks} ($\mathcal{F}^{\text{F}}$). \quad
The assembly of $\mathcal{F}^{\text{F}}$ is more modular, where the outputs of $K_{\text{F}}$ modules are simply weighted and aggregated.
When the $m^{\text{S}}$ module is not selected, $\mathcal{F}^{\text{F}}$ can be formalized as follows:
\begin{equation}
    \begin{aligned}
        &\mathcal{F}^{\text{F}}(\mathbf{u}) := \sum_{k\in \mathcal{K}_\text{F}} r^{\text{F}}_{k} \cdot m^{\text{F}}_k(\mathbf{u}), \\  
        &\mathbf{r}^{\text{F}} = \mathcal{R}^{\text{F}}(\mathbf{x})=\left(r^{\text{F}}_{1},\ldots,r^{\text{F}}_{k},\ldots, r^{\text{F}}_{K_F}\right).
    \end{aligned}
\end{equation}
When $m^{\text{S}}$ is chosen, likewise, only $K_{\text{F}} - 1$ \texttt{FFN}s form the $\mathcal{F}^{\text{F}}$.

\subsection{\name\ router (\texorpdfstring{$\mathcal{R}$}{})}
\label{sec:router}
In prior approaches, routing occurs as a one-step decision-making process within a layer. 
However, in \name\, which possesses a dynamically constructed computation graph, each decision is interdependent with the preceding ones, influencing the entire forward computation. 
Consequently, the router in \name\ necessitates an awareness of past decisions.
To model such dependency, we employ a gated recurrent unit (GRU,~\citep{cho-etal-2014-learning}) as the backbone of routers. 
Two routers in \name\ are identical in structure.
At each assembly step, the GRU in the $\mathcal{R}^\mathcal{X}$ maintains an $\mathbf{s}^{\mathcal{X}}_{h}$ as the hidden state of the GRU network. 
This state is recurrently updated as follows:
\begin{equation}
\begin{aligned}
    &\mathbf{s}^{\text{A}}_{h} = \texttt{GRU}^{\text{A}}(\mathbf{x}_{h}, \mathbf{s}^{\text{A}}_{h-1}), \\
    &\mathbf{s}^{\text{F}}_{h} = \texttt{GRU}^{\text{F}}(\mathbf{u}_{h}, \mathbf{s}^{\text{F}}_{h-1}).
\end{aligned}
\end{equation}
The weights assigned to each module by $\mathcal{R}^{\mathcal{X}}$ are computed as:
\begin{equation}
    \begin{aligned}
        &\mathbf{r}^{\mathcal{X}} = \mathbf{W}_{\mathcal{X}} \mathbf{s}^{\mathcal{X}}_{h},  \\
        &\mathbf{W}_{\mathcal{X}} \in \mathbb{R}^{(N_{\mathcal{X}}+1) \times d}, \quad \mathcal{X}\in \{\text{A, F}\}.
    \end{aligned}
\end{equation}
\subsection{Training approach}
\label{sec:optimize}
A straightforward approach is to pre-train an \name\ model initialized from scratch.
This approach, however, suffers from a degeneration issue, as the learned functions of modules become homogeneous, making router training challenging.
To address the issue, we propose a two-phase training approach. In the first phase, we pre-train a vanilla Transformer where modules acquire distinct functionalities. Then, in the second phase, we initialize the module set $\mathcal{M}$ with the pre-trained modules and initialize the routers from scratch. Subsequently, we continue training both modules and routers using the same data and objective as in the first phase. Through empirical studies, we find that the two-phase training method improves the specialization of module functionalities and accelerates router convergence.

\subsection{\name ~as a unified framework}
\label{sec:unify}

A compelling property of \name ~is that it unifies a wide range of Transformer-based dynamic computation allocation architectures. With specific configurations, layer-skip (e.g., early-exiting, mixture-of-depths, etc.), parameter sharing, and mixture-of-experts can be viewed as special cases.

\paragraph{Layer-skip.} The key idea is to skip layers according to certain criteria which can either be defined heuristically~\cite{liu2024accelerating} or learned from data~\cite{zeng2023learning,mod}.
Within the \name\ framework, layer-skip can be formulated as a special cluster of assembly functions $\phi$, namely:
\begin{equation}
\begin{aligned}
    \phi_{\text{layer-skip}}(\mathcal{M}, \mathcal{R}^{\mathcal{X}},\mathbf{x}_h)=\begin{cases}
        m_h^{\mathcal{X}} & \text{if}\ c_{\text{skip}}(h) = 1 \\
        m^S & \text{if}\ c_{\text{skip}}(h) = 0
    \end{cases},
\end{aligned}
\end{equation}
where $c_{\text{skip}}(\cdot)$ is the criterion that decides whether to skip the $h$-th layer or not. Note that the technique of early-exiting~\cite{graves2017adaptive,figurnov2017spatially} can be viewed as a special case of layer-skip, where once a layer is skipped, all subsequent layers will be skipped too.
\paragraph{Parameter sharing.} 
We consider parameter sharing that shares weights across modules and does not involve reparameterization techniques. Under this restriction, the sharing paradigm can be defined as a criterion function $c_{\text{share}}:i\mapsto j\ (j\le i)$, representing using the same weights for module $m_i^{\mathcal{X}}$ and module $m_j^{\mathcal{X}}$. Within \name,  parameter sharing can be formulated as:
\begin{equation}
\begin{aligned}
    \phi_{\text{parameter-sharing}}(\mathcal{M}, \mathcal{R}^{\mathcal{X}},\mathbf{x}_h)=m_{c_{\text{share}}(h)}^\mathcal{X}.
\end{aligned}
\end{equation}
\paragraph{Mixture-of-Experts.} MoE splits \texttt{FFN} into experts, and the experts are not shared across different layers. In MoE, routing is only performed on \texttt{FFN} modules, thus the computation of \texttt{MHA} is the same as that of a vanilla Transformer. The assembly function for MoE can be written as:

\begin{equation}
\begin{aligned}
    \phi_{\text{MoE}}(\mathcal{M}, \mathcal{R}^{\mathcal{X}},\mathbf{x}_h)= \begin{cases}
        m_h^\text{A} & \text{if}\ \mathcal{X}=\text{A}\\
        \phi_{\text{MoM}}(\mathcal{M}_h, \mathcal{R}^{\text{F}},\mathbf{x}_h) & \text{if}\ \mathcal{X}=\text{F}\\
    \end{cases},
\end{aligned}
\end{equation}
where $\mathcal{M}_h=\{m_{h,i}^{\text{F}}\}_{i=1}^{N_{\text{F}}}$ is the collection of experts for layer $h$.

\begin{figure*}[t]
    \centering
    \includegraphics[width=0.8\linewidth]{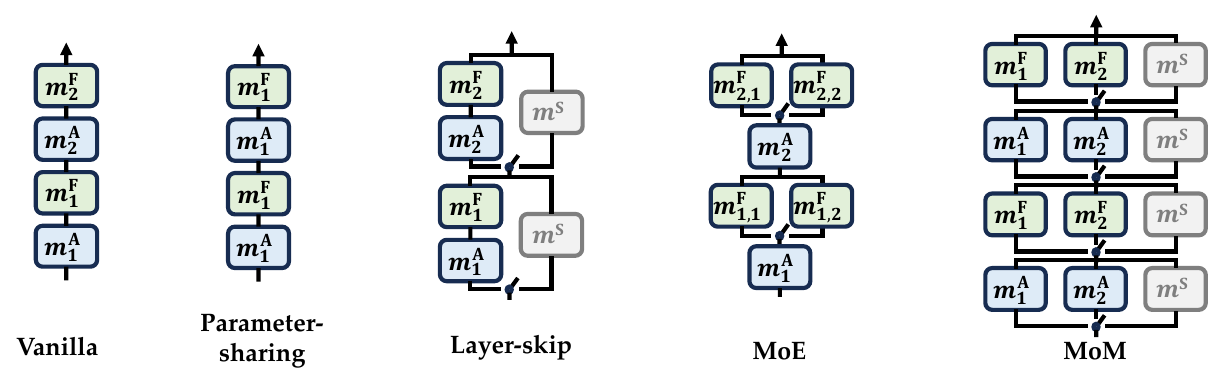}
    \caption{
    Visualization of forward computation of five models, where each consists of only two layers just for demonstration purposes.
    The switch icon symbolizes the selective execution of one (in Layer-skip) or more (in MoE and \name) subsequent computation pathways.}
    \label{fig:unify}
\end{figure*}

Figure~\ref{fig:unify} illustrates the forward computation process across different methods, offering an intuitive presentation of the versatility and universality in \name.

\section{Experiments}
\subsection{Experimental setup}
\label{sec:exp-setup}
We implement language models with \name\ since language modeling is a challenging task requiring both language understanding and generation ability, thereby effectively evaluating \name\ and baselines.
Below, we elaborate on the implementation details of \name, baselines, and evaluation setups.

\paragraph{Implementation details.} 
We conduct experiments on three model scales, which we denote as \name-small, \name-medium, and \name-large, respectively. These models contain 122M, 346M, and 774M parameters, respectively. Detailed configurations can be found in Appendix~\ref{apx:detail}.
The vanilla Transformers used for initializing \name\ are official GPT-2 checkpoints downloaded from HuggingFace\footnote{\url{https://huggingface.co/openai-community/gpt2}}.
$K$ and $H$ represent two hyper-parameters of \name. We denote a configuration where $K=a, H=b$ as K$a$H$b$. If the skip module is included in $\mathcal{M}$, we append the suffix S to K$a$H$b$.


In the two-phase training, we exploit OpenWebText~\citep{Gokaslan2019OpenWeb} as the pre-training dataset, and pre-process the data with the same pipeline as nanoGPT~\citep{nanoGPT}.
OpenWebText contains $9$ billion tokens after tokenization, from which $4$ million tokens are randomly sampled as the validation set.
The training sequence length for every input is $1,024$.
We set the learning rate to $1$e-$3$ with a warm-up ratio of 0.1 throughout the two phases, and do not use dropout.
All models are trained on 8 $\times$ A100 GPUs with a total batch size of 8 $\times$ 64. 
Two training phases require $20$k and $10$k optimization steps, respectively.

In practice, considering the large search space characterized by $H$ and $K$, we confine the architecture search space to a practical scale with a ``chunking'' strategy. An \name\ is divided into several chunks. Each chunk is independent and parameterized with identical $H$ and $K$. 
We present a detailed description and specific configuration in Appendix~\ref{apx:detail}, and an empirical analysis of the efficacy of chunking in Appendix~\ref{apx:chunk}.



\begin{table*}
    \centering
    \resizebox{\linewidth}{!}{
    \begin{tabular}{llrrrrrrr}
    \toprule
       \textbf{Methods} & \name\ Config & Parameter & Computation & Memory & Validation & Validation & GLUE & XSUM\\
        \textbf{$\phi$}& (K$a$H$b$) & Count & Cost (TFLOPs) & Cost (Gb) & Loss & Perplexity & (Average) & (Average) \\
    \midrule
    \midrule
    \multicolumn{9}{c}{\textit{small}}\\
    \midrule
         GPT2 & K1H4 & 122M  & 2.92 & 2.98 & 3.10 & 22.22 & 75.32 & \underline{14.26} \\
         MoD & K1H4S & 122M & - & - & 3.22 & 25.11 & 72.24 & 9.71 \\
         MoE & K2H4 & 283M & 3.81 (\textcolor{red}{+30.5\%}) & 2.98 (\textcolor{red}{+0.0\%}) & 3.07 & 21.18 & 77.25 & 14.18 \\
         MoE (share) & K2H4 & 122M & 3.81 (\textcolor{red}{+30.5\%}) & 2.98 (\textcolor{red}{+0.0\%}) & 3.14 & 23.41 & 75.82 & 14.15 \\
         $\text{\name}_{\text{E}}$ & K3H1S & 122M & 2.45 (\textcolor{blue}{-16.1\%}) & 2.45 (\textcolor{blue}{-17.8\%}) & 3.16 & 23.59 & 75.92 & 14.17 \\
         $\text{\name}_{\text{I}}$ & K3H2S & 122M & 3.49 (\textcolor{red}{+19.5\%}) & 2.63 (\textcolor{blue}{-11.7\%}) & \underline{3.03} & \underline{20.79} & \underline{77.81} & 14.24 \\
         $\text{\name}_{\text{P}}$ & K2H6S & 122M & 5.04 (\textcolor{red}{+72.6\%}) & 3.34 (\textcolor{red}{+12.1\%}) & \textbf{2.98} & \textbf{19.59} & \textbf{78.22} & \textbf{15.19} \\
    \midrule
    \multicolumn{9}{c}{\textit{medium}}\\
    \midrule
         GPT2 & K1H4 & 346M & 8.28 & 4.74 & 2.81 & 16.69 & 80.49 & 18.14\\
         MoD & K1H4S & 346M  & - & - & 2.99 & 19.82 & 76.17 & 14.81 \\
         MoE & K2H4 & 921M & 11.37 (\textcolor{red}{+37.3\%}) & 4.74 (\textcolor{red}{+0.0\%}) & 2.80 & 16.53 & 80.47 & 17.75 \\
         MoE (share) & K2H4 & 346M & 11.37 (\textcolor{red}{+37.3\%}) & 4.74 (\textcolor{red}{+0.0\%}) & 2.82 & 16.81 & 80.35 & 17.59 \\
         $\text{\name}_{\text{E}}$ & K3H1S & 346M & 6.80 (\textcolor{blue}{-17.9\%}) & 3.33 (\textcolor{blue}{-29.7\%}) & 2.83 & 16.91 & 80.41 & 17.11\\
         $\text{\name}_{\text{I}}$ & K3H2S & 346M & 10.20 (\textcolor{red}{+23.2\%}) & 3.80 (\textcolor{blue}{-19.8\%}) & \underline{2.77} & \underline{15.89} & \underline{81.03} & \underline{18.66}\\
         $\text{\name}_{\text{P}}$ & K2H6S & 346M & 16.23 (\textcolor{red}{+96.0\%}) & 5.69 (\textcolor{red}{+20.0\%}) & \textbf{2.72} & \textbf{15.18} & \textbf{81.93} & \textbf{19.30}\\
    \midrule
    \multicolumn{9}{c}{\textit{large}}\\
    \midrule
         GPT2 & K1H4 & 774M  & 17.76 & 7.20 & 2.66 & 14.33 & 84.47 & 20.35 \\
         MoD & K1H4S & 774M & - & - & 2.81 & 16.62 & 81.49 & 18.62 \\
         MoE & K2H4 & 2100M & 25.43 (\textcolor{red}{+43.2\%}) & 7.20 (\textcolor{red}{+0.0\%}) & \underline{2.64} & 14.17 & 84.43 & 20.63 \\
         MoE (share) & K2H4 & 774M & 25.43 (\textcolor{red}{+43.2\%}) & 7.20 (\textcolor{red}{+0.0\%}) & 2.65 & 14.22 & 83.83 & 20.39 \\
         $\text{\name}_{\text{E}}$ & K3H1S & 774M  & 14.84 (\textcolor{blue}{-16.4\%}) & 4.13 (\textcolor{blue}{-42.6\%}) & 2.66 & 14.50 & 83.39 & 20.46 \\
         $\text{\name}_{\text{I}}$ & K3H2S & 774M & 20.31 (\textcolor{red}{+14.5\%}) & 5.15 (\textcolor{blue}{-28.5\%}) & \underline{2.64} & \underline{13.92} & \underline{84.49} & \underline{21.73} \\
         $\text{\name}_{\text{P}}$ & K2H6S & 774M & 36.07 (\textcolor{red}{+103.1\%}) & 9.24 (\textcolor{red}{+28.3\%}) & \textbf{2.60} & \textbf{13.21} & \textbf{85.90} & \textbf{22.36} \\
    \bottomrule
    \end{tabular}}
    \caption{Comprehensive comparison between \name s and baselines. 
    We highlight the best results in bold and underline the second-best results.
    Appendix~\ref{apx:glue} includes the detailed performance on GLUE and XSUM.}
    \label{tab:main}
\end{table*}

\paragraph{Baselines.} In addition to the vanilla Transformer model~\citep{Radford2019LanguageMA}, the following models (or methods) are also employed as baselines: 
(1) \textbf{MoD}~\citep{mod}: a layer-skip method proposed recently that dynamically routes around Transformer blocks.\footnote{As official code is unavailable until the submission, we follow the paper to implement MoD ourselves.} (2) \textbf{MoE}: the mixture-of-experts architecture utilized by  Mixtral. We implement the model with the open-sourced code.\footnote{\url{https://github.com/huggingface/transformers/blob/v4.36.1/src/transformers/models/mixtral/modeling_mixtral.py}} (3) \textbf{MoE (share)}: a variant of MoE in which all layers share the same set of experts. We involve this model as a baseline because unlike the standard MoE that has more parameters, MoE-share has the same number of parameters with the vanilla Transformer model, making the comparison more fair.  Moreover, it also sheds light on how well the MoE architecture can utilize a fixed budget of modules. 

Note that all the above-mentioned methods are special cases within the \name\ framework with various configurations. 
Furthermore, we explore a wide range of \name\ instances defined by K$a$H$b$S, where $a\le 4$ and $b\le 6$. 
We examine all \name\ instances within the search space (as detailed in \S\ref{sec:config}), and spotlight three distinct models for comparison against other baselines:

$\bullet$ \textbf{$\text{\name}_\text{P}$} (K2H6S) represents a \textbf{P}erformant \name\ model after tuning $K$ and $H$. 

$\bullet$ \textbf{$\text{\name}_\text{E}$} (K3H1S) significantly enhances \textbf{E}fficiency compared to vanilla Transformers, while maintaining acceptable performance.

$\bullet$ \textbf{$\text{\name}_\text{I}$} (K3H2S) serves as a midpoint in configurations between the two preceding models. This model is positioned between $\text{\name}_\text{E}$ and $\text{\name}_\text{P}$. We aim to highlight performance and efficiency \textbf{I}nterpolation as feature of \name\ through configuration interpolation.

\paragraph{Evaluation settings.} We employ GLUE benchmark~\citep{wang2018glue} to evaluate the language understanding ability and XSUM~\citep{xsum-emnlp} to evaluate the text generation ability. All models are fine-tuned with a learning rate of $2$e-$5$. The sequence is $128$ for GLUE and $1024$ for XSUM. For smaller GLUE sub-datasets (CoLA, STS-B, MRPC, and RTE), we set the batch size to 32 and train for 3 epochs. For larger datasets (MNLI, QNLI, QQP, and SST-2), we utilize a batch size of 64 and perform training for a total of $8,000$ gradient steps. For XSUM, we set the batch size to 64 and train for 3 epochs. 
For efficiency evaluation, we report inference TFLOPs and memory usage. 
TFLOPs are calculated using DeepSpeed FLOPs profiler~\citep{flops-profiler} and memory consumption is calculated with PyTorch toolkits~\citep{PyTorch-Profiler}.

\subsection{Main results} 
\label{sec:main}
Table~\ref{tab:main} reports the evaluation results. 
Our analysis yields the following conclusions:

\paragraph{\name\ unleashes the potential of Transformers and our initial motivation is confirmed.}
When maintaining the number of parameters, $\text{\name}_{\text{P}}$ is characterized by the deepest computation graph ($H$).
Across all model scales, $\text{\name}_{\text{P}}$ consistently outperforms all baselines on both GLUE and XSUM by significant margins.
The enhanced performance of $\text{\name}_{\text{P}}$ validates our initial motivations: (1) the traditional depth-ordered layer organization is sub-optimal; 
(2) improvements can be realized through two key modifications to the computation graph, including \textit{dynamic module organization} and \textit{improved parameter utilization}.

$\text{\name}_{\text{E}}$ is characterized by its minimum depth ($H$).
By strategically selecting appropriate modules at each assembly step, $\text{\name}_{\text{E}}$ strives to reduce memory and computation costs while maintaining performance.
Although $\text{\name}_{\text{E}}$ is slightly surpassed by a vanilla Transformer, it outperforms MoD, another efficiency-driven method by large margins.

Besides, we observe that $\text{\name}_{\text{I}}$ archives a decent performance by slightly outperforming vanilla GPT-2.
Comparing to vanilla GPT-2, $\text{\name}_{\text{I}}$ consumes no more than 25\% extra computation but save at least 11.7\% memory across all scales, indicating that $\text{\name}_{\text{I}}$ achieves a good balance between performance and efficiency. 



\begin{figure}[h]
    \centering
    \includegraphics[width=0.3\linewidth]{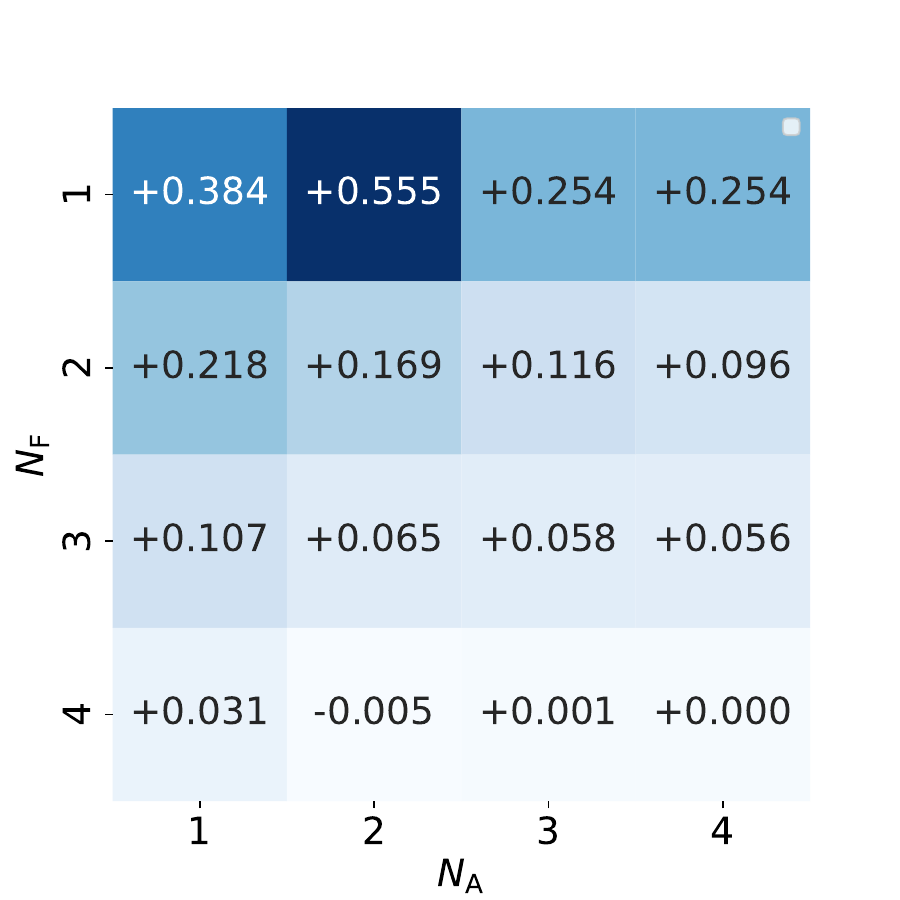}
    \caption{How validation loss varies with respect to $N_{\text{A}}$ and $N_{\text{F}}$, comparing to \name\ (medium) with $N_{\text{A}}=N_{\text{F}}=4$.} 
    \label{fig:heatmap}
\end{figure}
\paragraph{\name\ models provide insights into the over-parameterization issue.} With GPT-2-medium as initialization,
we develop a series of \name\ models, each defined by different pairs of ($N_{\text{A}}, N_{\text{F}}$), with both values not exceeding $4$.
We assess the validation loss increase for each model relative to the benchmark model where $N_{\text{A}}=4$ and $N_{\text{F}}=4$, as illustrated in Figure~\ref{fig:heatmap}.
In this experiment, we set $K$ equal to the number of modules to make full use of the module parameters.
Interestingly, the \texttt{FFN} and \texttt{MHA} modules exhibit different degrees of redundancy.
Specifically, when $N_\text{F}$ is fixed and the number of \texttt{MHA}s is gradually reduced, a significant increase in loss is not observed until $N_\text{A}$ is reduced from $2$ to $1$, suggesting considerable redundancy in the \texttt{MHA}s of Transformers.
In contrast, when fixing $N_\text{A}$ and reducing the number of \texttt{FFN}s gradually, each time of removing an \texttt{FFN} leads to evident loss increase, indicating \texttt{FFN}s are less over-parameterized.
These quantitative findings align with previous research suggesting that the parameterization of attention can be simplified to enhance efficiency while maintaining performance~\citep{deepseekai2024deepseekv2,shazeer2019fast}. 

\paragraph{As the parameter size scales up, \name\ models enjoy consistent gains in both performance and efficiency.}
When we look into the difference across different scales, we observe that (1) the performance gain of \name\ is stable; (2) $\text{\name}_{\text{E}}$-medium and  $\text{\name}_{\text{E}}$-large exhibit more significant reductions in resource costs comparing to $\text{\name}_{\text{E}}$-small.
These observations across different scales reinforce our previous motivation: Transformers are over-parameterized, which becomes more evident as the model size increases. 

\subsection{Insights from hyper-parameter search}
\label{sec:config}
\begin{figure}[h]
\vspace{-4mm}
    \centering
    \includegraphics[width=0.5\linewidth]{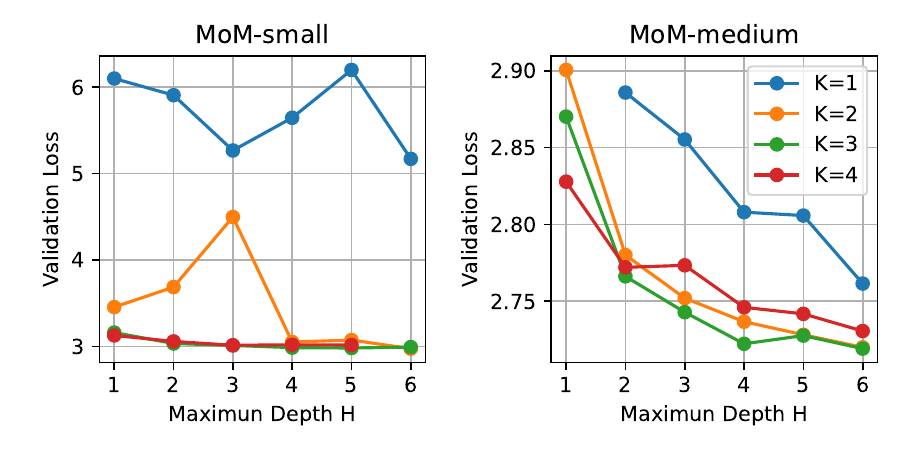}
    \caption{Validation loss for \name-small and \name-medium under different settings of $K$ and $H$.}
    \label{fig:hk}
    \vspace{-2mm}
\end{figure}

Figure~\ref{fig:hk} shows how the validation loss for \name-small and \name-medium varies with respect to different settings of $K$ and $H$ ($K\in\{1,2,3,4\}$, $H\in\{1,2,3,4,5,6\}$). From this experiment, we have the following observations and insights: (1) allowing more modules to be assembled at each step (i.e., larger $K$) and more rounds of assembling actions (i.e., larger $H$) generally leads to better performance, indicating that \textit{Transformer-based models benefit from a larger computation graph even if the parameter size remains the same}. However, (2) the benefits of increasing $K$ and $H$ become marginal when $K > 2$ and $H > 1$. 
Comparing K3H1 to K2H6, we can see that the validation loss is comparable, while K3H1 performs slightly worse on downstream tasks as discussed in \S\ref{sec:main}. However, K3H2 improves efficiency by flattening the depth, making it a good choice that balances performance and efficiency.
Flattening modules from different depths to the same depth cancels computation dependencies of each other. This characteristic brings an extra benefit because the computation of modules from the same depth can be parallelized.
This technique has been validated and adopted in MoE applications~\citep{switch-trans,lepikhin2021gshard} (called expert parallelism) and can be easily extended to further accelerate \name\ (K3H2).

\begin{table}
    \centering

    \begin{tabular}{lllcc}
    \toprule
       & Phase 1 & Phase 2 & Val. Loss & PPL.\\ \midrule
    (1) & MoM & - & 2.85 & 17.26 \\
    (2) & Vanilla & - & 2.81 & 16.69 \\
    (3) & MoM & MoM & 2.78 & 16.20 \\
    (4) & Vanilla & MoM & \bf{2.72} & \bf{15.18} \\
    (5) & Vanilla & MoM-same & 3.13 & 22.87 \\
    \bottomrule
    \end{tabular}
    \caption{Performance comparison under different training setups. When training \name\ from scratch, we set the total gradient steps of phase 1 to 20k. The total steps of phase 2 for all ablations are 10k.}
    \label{tab:pretrain_phase}
\end{table}

\subsection{Impact of two-phase training}
\label{exp:two-phase-pretrain}
    %
We investigate how the two-phase training strategy influences model performance using $\text{\name}_\text{P}$-medium as case studies. Specifically, we consider the following training strategies: (1) training \name\ from scratch with 20k steps; (2) training vanilla Transformer with 20k steps; (3) training \name\ from scratch with 30k steps; (4) $\text{\name}_\text{P}$-medium that is trained with the two-phase strategy; and (5) training \name\ with the two-phase strategy but in the second phase, \texttt{MHA} modules and \texttt{FFN} modules are forced to be identical, respectively. Table~\ref{tab:pretrain_phase} shows the results. First, comparing (1) with (2), we find that the absence of weight initialization compromises the training quality of \name, making it worse than the vanilla Transformer, underscoring the importance of initializing module weights with a well-trained vanilla Transformer model for \name. A plausible explanation for the results is that training without warm-up leads to homogeneous modules that hurt the convergence of the routers (as illustrated in Appendix~\ref{apx:router}). The explanation is further justified by comparison between (4) and (5), as when we force all modules to be identical, the performance of \name\ also drops dramatically even with the two-phase training strategy. As expected, training with more steps can enhance performance (cf., comparison between (1) and (3)), but the two-phase strategy is still the better choice when we compare (3) with (4).


\section{Conclusions}
In this work, we propose \fullname\ (\name), a novel architecture that reinvents transformers as a collection of individual modules and the dynamic assembly process conducted with these modules. This novel view offers us an opportunity to explore a wide range of different configurations of model architecture and unify a series of transformer variants. With exhaustive experiments, we not only validate the effectiveness of \name\ by both significant efficiency and performance gains but also reach new insights about Transformers.

\section*{Limitations}
Our current design of the router still has room for improvement. Unlike MoE wherein the router makes one-time decisions about which experts to select, the router of \name\ is responsible for conducting multi-step decision-making. In this scenario, instructing the router to make correct decisions continuously is a hard problem since the decision space grows exponentially with the increase of assembly steps. The current implementation has not considered this question and has not explicitly encouraged or discouraged the router to make some choices, thus, we are not sure whether the learned routing decisions are optimal or not. In the future, we will explore using techniques like reinforcement learning or neural architecture search to design more sophisticated routers.

\bibliography{my}
\bibliographystyle{iclr2024_conference}

\appendix
\section{More implementation details}
\label{apx:detail}
Table~\ref{tab:scale_config} lists the configurations of \name-small/medium/large.

\begin{table*}[h]
    \centering
    \begin{tabular}{cccc}
    \toprule
         & \textbf{\name-small} & \textbf{\name-medium}  & \textbf{\name-large}\\
         \midrule
         Initialization model & GPT2-small & GPT2-medium & GPT2-large \\
         Hidden size & 768 & 1024 & 1280 \\
         Total number of \texttt{FFN}/\texttt{MHA} modules & 12 & 24 & 36 \\
         Number of attention heads & 12 & 16 & 20 \\
         Max sequence length & 1024 & 1024 & 1024 \\
         Vocabulary size & 50257 & 50257 & 50257 \\
    \bottomrule
    \end{tabular}
    \caption{Model configurations for \name-small/medium/large.}
    \label{tab:scale_config}
\end{table*}

In practice, we segment \name\ into equally-sized chunks, each containing $4$ \texttt{MHA} modules and $4$ \texttt{FFN} modules, namely $N=4$. Within each chunk, we execute the \name\ assembly process as presented in \S\ref{sec:method}. 
We restrict the search space of each chunk by setting $K\le 4$ and $H\le 6$, which results in $4\times6=24$ combinations in total.

Then we elaborate on the initialization of chunked \name. Taking a 8-layer vanilla transformer as an example, the architecture is sliced as: the bottom/top 2 layers remain the same, and modules in the middle 4 layers form a \name\ block, wherein we conduct the iterative assembly process. We denote this chunking strategy as \texttt{[1-1-4-1-1]} where ``\texttt{1}" represents the standard Transformer block and ``\texttt{4}" represents a chunk whose $N$ equals to 4. Empirically, we find this setting to be stable across various choices of K$a$H$b$. A detailed experimental analysis of different chunking strategies can be found in Appendix~\ref{apx:chunk}. Similarly, for \name-small, \name-medium and \name-large, we use the chunking strategies of \texttt{[1-1-4-1-4-1]}, \texttt{[1-1-1-4-1-4-1-4-1-4-1-1]}, and \texttt{[1-4-1-4-1-4-1-4-1-4-1-4-1-4-1]}, respectively.

\section{Chunks}
\label{apx:chunk}

\begin{table}[h]
    \centering
    \begin{tabular}{lll}
    \toprule
      Chunking Strategies & MoM Config & Val. Loss \\ \midrule
      \texttt{[1-1-4-1-1]} & K1H4S & 3.27\\
      \texttt{[1-1-4-1-1]} & K2H4S & 3.22\\
      \texttt{[1-6-1]} & K1H6S & 3.45\\
      \texttt{[1-6-1]} & K2H6S & 3.21\\
      \texttt{[8]} & K1H8S & 4.63\\
      \texttt{[8]} & K2H8S & 3.23\\
      \texttt{[4-4]} & K1H4S & 5.59\\
      \texttt{[4-4]} & K2H4S & 3.22\\
    \bottomrule
    \end{tabular}
    \caption{Applying different chunking strategies on an 8-layer MoM. These models follow the same two-phase training procedure and the total training steps of the second phase is 5k.}
    \label{tab:chunk}
\end{table}
In this section, we study the impact of different chunking strategies on \name\ performance. This experiment is conducted on an 8-layer \name.
Except for \texttt{[1-1-4-1-1]}, we include several alternatives: \texttt{[4-4]} (two successive \name\ blocks with $N=4$), \texttt{[1-6-1]} (the top and bottom one layer are kept unchanged, and modules in the middle 6 layers form an \name\ with $N=6$), and \texttt{[8]} (all modules form a big \name\ with $N=8$).
Table~\ref{tab:chunk} shows the results of different chunking strategies. When $K=1$, strategies other than \texttt{[1-1-4-1-1]} exhibit unstable training curves and bad performance.
This is because the routers need to make multi-step decisions in the search space. 
A larger search space (the increase of $N$) and more assembly steps (the increase of $H$) all lead to a harder task for the routers to find the correct path in the search space.
Things are much better when $K=2$, where all strategies converge quite well, indicating that we can develop a full \name\ architecture (i.e.,\texttt{[8]}) without the chunking strategy. In practice, we still apply the chunking strategy (i.e., \texttt{[1-1-4-1-1]}), because the strategy allows us to vary $K$ in a larger range, and thus we can study \name\ with more configurations. 
The experiment demonstrates the necessity of manually restricting the search space of \name\ so that the decision-making burden for the routers would be relieved.

\section{Detailed downstream evaluation results}
\label{apx:glue}
Table \ref{tab:glue} presents the evaluation results for each sub-task of GLUE across different models and Table \ref{tab:xsum} presents the results on XSUM with respect to other ROUGE metrics.

\begin{table*}
    \centering
    \resizebox{\linewidth}{!}{
    \begin{tabular}{lccccccccc}
    \toprule
        Method & SST-2 & COLA & MRPC & QQP & QNLI & RTE & MNLI-(m/mm) & STS-B & average  \\
    \midrule
    \midrule
        \multicolumn{10}{c}{\textit{small}} \\
    \midrule
        GPT2 & 92.09 & 26.27 & 85.11 & 83.09 & 87.55 & 62.45 & 79.90/78.74 & 82.67 & 75.32 \\
        MoD & 87.73 & 21.66 & 81.43 & 82.01 & 81.79 & 64.62 & 75.96/75.22 & 79.76 & 72.24 \\
        MoE & 89.68 & 45.87 & 82.05 & 84.09 & 85.94 & 65.52 & 79.46/78.74 & 83.87 & 77.25 \\
        MoE (share) & 89.53 & 37.29 & 82.01 & 83.49 & 85.21 & 64.94 & 78.49/77.99 & 83.43 & 75.82 \\
        $\text{\name}_{\text{E}}$ & 90.83 & 31.83 & 82.92 & 83.49 & 84.79 & 69.68 & 78.52/77.40 & 83.84 & 75.92 \\
        $\text{\name}_{\text{I}}$ & 90.02 & 47.21 & 82.90 & 84.07 & 85.34 & 66.43 & 79.82/79.72 & 84.79 & 77.81 \\
        $\text{\name}_{\text{P}}$ & 91.40 & 46.16 & 82.68 & 84.62 & 86.49 & 67.51 & 80.85/80.18 & 84.06 & 78.22 \\
    \midrule
        \multicolumn{10}{c}{\textit{medium}} \\
    \midrule
        GPT2 & 94.15 & 48.18 & 86.00 & 85.94 & 90.41 & 64.98 & 84.02/83.92 & 86.77 & 80.49 \\
        MoD & 89.45 & 37.43 & 84.97 & 84.21 & 84.88 & 64.26 & 78.94/77.94 & 83.47 & 76.17 \\
        MoE & 91.86 & 49.20 & 85.51 & 86.39 & 89.09 & 68.12 & 84.44/83.49 & 86.17 & 80.47 \\
        MoE (share) & 92.78 & 49.44 & 84.56 & 86.40 & 89.35 & 66.79 & 84.00/83.39 & 86.42 & 80.35 \\
        $\text{\name}_{\text{E}}$ & 91.40 & 48.48 & 87.48 & 86.92 & 89.11 & 67.51 & 83.73/83.01 & 86.06 & 80.41 \\
        $\text{\name}_{\text{I}}$ & 92.46 & 50.19 & 87.31 & 86.43 & 89.35 & 68.94 & 84.44/83.46 & 86.66 & 81.03 \\
        $\text{\name}_{\text{P}}$ & 92.88 & 53.61 & 87.64 & 86.64 & 89.75 & 71.06 & 84.69/83.98 & 87.14 & 81.93 \\
    \midrule
        \multicolumn{10}{c}{\textit{large}} \\
    \midrule
        GPT2 & 94.15 & 60.04 & 88.74 & 87.88 & 91.89 & 75.45 & 86.80/85.98 & 89.30 & 84.47 \\
        MoD & 91.51 & 52.86 & 87.29 & 87.20 & 89.57 & 67.87 & 85.07/84.38 & 87.66 & 81.49 \\
        MoE & 94.32 & 61.19 & 88.54 & 88.36 & 92.04 & 71.68 & 87.38/86.94 & 89.45 & 84.43 \\
        MoE (share) & 93.88 & 61.43 & 88.12 & 88.17 & 91.56 & 68.31 & 87.01/86.73 & 89.28 & 83.83 \\
        $\text{\name}_{\text{E}}$ & 93.64 & 59.26 & 88.25 & 87.59 & 91.29 & 72.23 & 85.20/84.78 & 88.26 & 83.39 \\
        $\text{\name}_{\text{I}}$ & 93.69 & 62.25 & 89.19 & 88.12 & 92.36 & 74.98 & 87.22/86.78 & 89.90 & 84.94 \\
        $\text{\name}_{\text{P}}$ & 94.42 & 64.49 & 89.56 & 88.68 & 92.94 & 77.69 & 87.68/86.98 & 90.70 & 85.90 \\

    \bottomrule
    \end{tabular}}
    \caption{Detailed evaluation on the GLUE benchmark. We follow the previous evaluation setting~\cite{radford2018improving}, for SST-2, QNLI, RTE, and MNLI, we report accuracy as the metric. For MRPC and QQP, we report the F1 score. For STS-b, we report the combined score of Pearson correlation and Spearman correlation.}
    \label{tab:glue}
\end{table*}

\begin{table*}
    \centering
    \begin{tabular}{lrrrr}
    \toprule
       Method & ROUGE-1 & ROUGE-2 & ROUGE-L & ROUGE-AVG \\
    \midrule
    \midrule
    \multicolumn{5}{c}{\textit{small}}\\
    \midrule
         GPT2 & 20.8 & 5.05 & 16.92 & 14.26\\
         MoD & 14.45 & 2.89 & 11.80 & 9.71 \\
         MoE & 20.56 & 5.26 & 16.71 & 14.18 \\
         MoE (share) & 20.64 & 5.17 & 16.64 & 14.15 \\
         $\text{\name}_{\text{E}}$ & 20.62 & 4.98 & 16.91 & 14.17\\
         $\text{\name}_{\text{I}}$ & 20.50 & 5.31 & 16.90 & 14.24 \\
         $\text{\name}_{\text{P}}$ & 21.73 & 6.14 & 17.72 & 15.19 \\
    \midrule
    \multicolumn{5}{c}{\textit{medium}}\\
    \midrule
         GPT2 & 25.07 & 8.35 & 21.00 & 18.14\\
         MoD & 20.89 & 6.04 & 17.52 & 14.81 \\
         MoE & 24.58 & 8.20 & 20.49 & 17.75 \\
         MoE (share) & 24.36 & 8.19 & 20.22 & 17.59 \\
         $\text{\name}_{\text{E}}$ & 24.56 & 7.04 & 19.72 & 17.11 \\
         $\text{\name}_{\text{I}}$ & 26.29 & 8.21 & 21.47 & 18.66 \\
         $\text{\name}_{\text{P}}$ & 26.61 & 9.05 & 22.24 & 19.30 \\
    \midrule
    \multicolumn{5}{c}{\textit{large}}\\
    \midrule
         GPT2 & 28.16 & 9.92 & 22.98 & 20.35\\
         MoD & 26.08 & 8.38 & 21.41 & 18.62 \\
         MoE & 28.54 & 9.89 & 23.47 & 20.63 \\
         MoE (share) & 28.47 & 9.47 & 23.23 & 20.39 \\
         $\text{\name}_{\text{E}}$ & 28.48 & 9.65 & 23.24 & 20.46 \\
         $\text{\name}_{\text{I}}$ & 30.91 & 10.21 & 24.08 & 21.73 \\
         $\text{\name}_{\text{P}}$ & 31.38 & 10.77 & 24.93 & 22.36 \\
    \bottomrule
    \end{tabular}
    \caption{Detailed evaluation on the XSUM dataset. ROUGE is employed as the evaluation metrics.}
    \label{tab:xsum}
\end{table*}

\section{Router analysis}
\label{apx:router}
\subsection{Architecture design}
\begin{figure}[h]
\vspace{10mm}
    \centering  
    \includegraphics[width=0.5\linewidth]{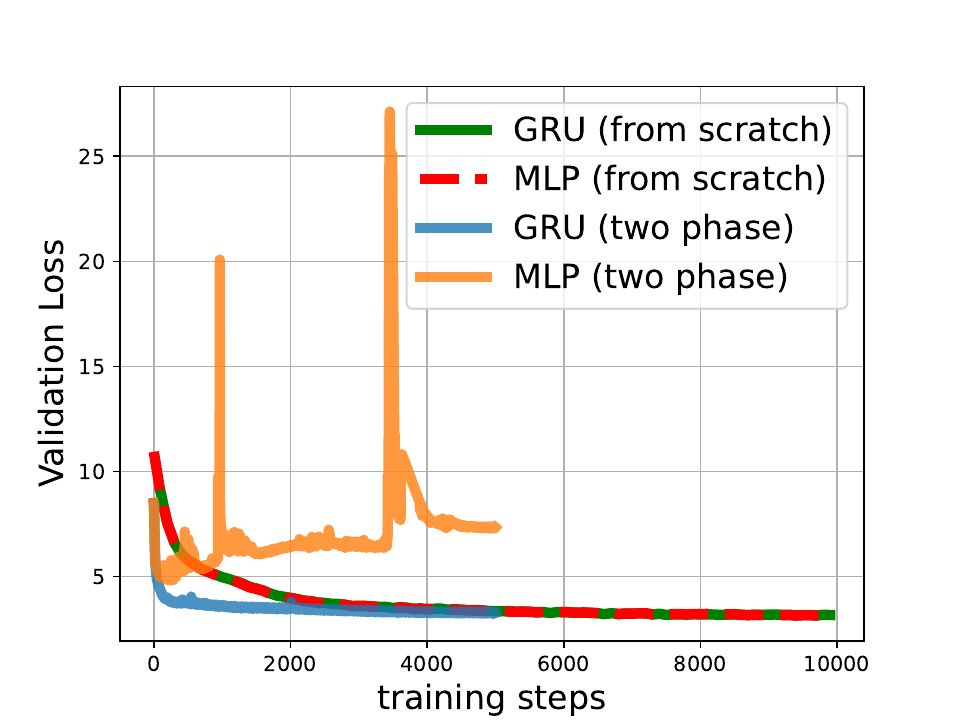}
    \caption{Training curves of \name-small (K1H4) with \{GRU, MLP\} routers.}
    \label{fig:training_curve}
\end{figure}
We study the implementation of a key component in \name: the routers.
We substitute the GRU within the router with a simple two-layer MLP, eliminating the interaction among router decision states.
Our exploration of the router's impact involves two setups: (a) initializing \name\ from scratch, and (b) employing the two-phase training approach. 
Here are some intriguing results.
As depicted in Figure~\ref{fig:training_curve}, when initializing from scratch, both router structures exhibit nearly identical loss curves.
However, under setting (b), training \name\ with the MLP router becomes unstable marked by spikes in gradient magnitude throughout the training.
This instability suggests the router's inability to establish a consistent assembly plan for tokens.
When initializing from scratch, the required capability may not be learned efficiently by these modules, as they often develop homogeneous functionalities that waste the parameters.
Conversely, in setting (b), where modules are initialized with specialized functions, the optimization progresses smoothly and converges quickly.

\subsection{Learned router patterns}
\begin{figure}[h]
    \centering
    \includegraphics[width=0.3\linewidth]{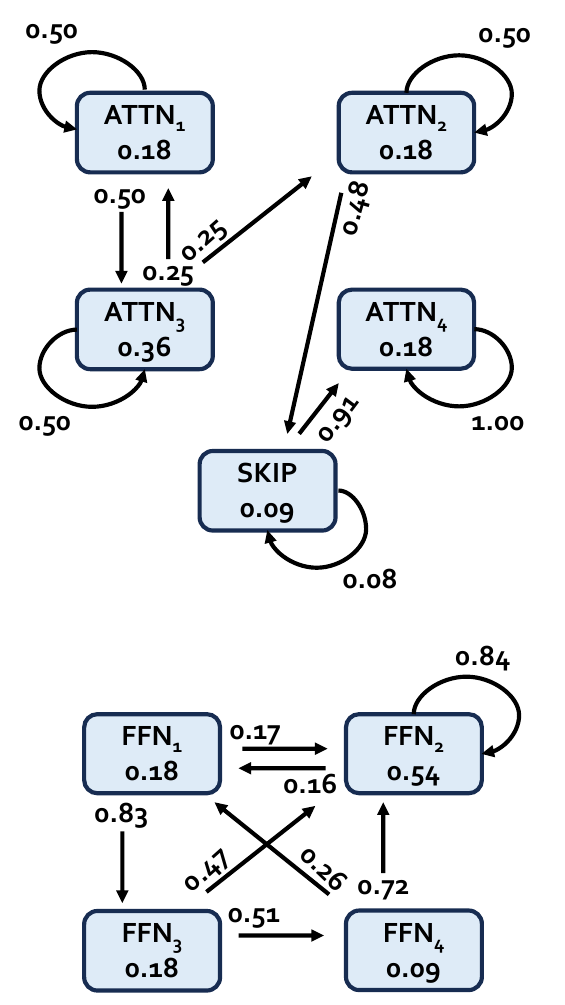}
    \caption{Routing patterns of \name-medium.}
    \label{fig:visualize}
\end{figure}
We are curious whether the router follows specific patterns when choosing and assembling modules. 
We visualize the transition probabilities between modules (Figure~\ref{fig:visualize}) to answer this. The first observation is that the router does not degrade into simply memorizing the original shallow-to-deep order but jumps across modules as expected. For example, a common routing path in Figure~\ref{fig:visualize} is $(2,2,1,3,4,2)$ for \texttt{FFN} modules and $(3,1,3,2,S,4)$ for \texttt{MHA} modules (number represents the module index). 
Another observation is that the loads of different modules are imbalanced. 
In some cases, specific modules are hardly used. 
Unlike MoE, which uses an auxiliary loss to balance the loads across different experts~\citep{switch-trans}, we do not see a positive effect by adding a balance loss to \name. 
Adding regularization alleviates the imbalance issue at the cost of performance degradation (by increasing validation perplexity by 1.8 points). 
We posit an intuitive explanation: within the language model framework, tasks that can be decomposed into numerous sub-tasks may exhibit various levels of difficulty.
Consequently, some sub-tasks necessitate more engagement of modules with specific processing capabilities, thus contributing to the observed imbalance.

\end{document}